\documentclass{INTERSPEECH2023}

\interspeechcameraready 

\usepackage{wasysym}
\usepackage{hyperref}

\title{Vistaar: Diverse Benchmarks and Training Sets for Indian Language ASR}
\name{Kaushal Santosh Bhogale$^1$*, Sai Sundaresan$^2$*\dag \thanks{* The first two authors have contributed equally.}\thanks{\dag Work was done when Sai was at AI4Bharat}, Abhigyan Raman$^3$, Tahir Javed$^4$, Mitesh M. Khapra$^5$, Pratyush Kumar$^6$}
\address{
$^{1,2,3,4,5,6}$AI4Bharat, India \\
  $^{1,4,5,6}$Indian Institute of Technology Madras, India\\
  $^{2}$Indian Institute of Technology Kharagpur, India\\
  $^{6}$Microsoft, India
  }
\email{cs22d006@cse.iitm.ac.in, {tahirjmakhdoomi,saisundaresan01,ramanabhigyan}@gmail.com, {miteshk,pratyush}@cse.iitm.ac.in}
\begin{document}

\maketitle
 
\begin{abstract}
Improving ASR systems is necessary to make new LLM-based use-cases accessible to people across the globe.
In this paper, we focus on Indian languages, and make the case that diverse benchmarks are required to evaluate and improve ASR systems for Indian languages. 
To address this, we collate Vistaar as a set of 59 benchmarks across various language and domain combinations, on which we evaluate 3 publicly available ASR systems and 2 commercial systems.
We also train IndicWhisper models by fine-tuning the Whisper models on publicly available training datasets across 12 Indian languages totalling to 10.7K hours. 
We show that IndicWhisper significantly improves on considered ASR systems on the Vistaar benchmark.
Indeed, IndicWhisper has the lowest WER in 39 out of the 59 benchmarks, with an average reduction of 4.1 WER. 
We open-source all datasets, code and models - \url{https://github.com/AI4Bharat/vistaar}

 
\end{abstract}

\section{Introduction}

Large Language Models such as GPT3\cite{brown2020language} have demonstrated emergent behaviors which unlock new use-cases for deployment of language AI. 
For instance, an Indian villager could interact with a LLM-based system to learn about rights and entitlements on existing government schemes.
To make such use-cases accessible to a large population, it is essential to make them voice-enabled, which requires support for speech recognition models across languages. 
Specifically, for Indian languages, given the large linguistic diversity (60 languages with over a million speakers) and the large print illiterate population (over 300 million people \cite{census}), accurate ASR systems have significant societal impact.

Improvement of Machine Learing systems like ASR are dependent on various choices of training architectures, algorithms, datasets, and metrics. 
Systematic evaluation of these choices is critically dependent on representative benchmarks. 
Several works, over many years, have contributed publicly available benchmarks of specific languages and specific domains/types of data. 
For instance, Gramvaani \cite{gramvaani}, a Hindi dataset, contains data from agriculture domain; Vakyasancayah \cite{vakyasancayah}, a Sanskrit dataset, contains data from literature domain.
Often these benchmarks are accompanied by leaderboards which earmark performance of different ASR systems.
We make the case that such narrow comparison incentivizes model optimization to over-fit for the benchmark's characteristics.
For instance, tuning of language models to the dev set of benchmarks have been shown to reduce WER scores by over 6.5 points averaged across 8 languages \cite{indicwav2vec}.
This suggests that narrow benchmarking may incentivize such over-fitting and thereby preclude more robust comparison of models.

\begin{figure}[t!]
    \centering
    \includegraphics[width=0.8\columnwidth]{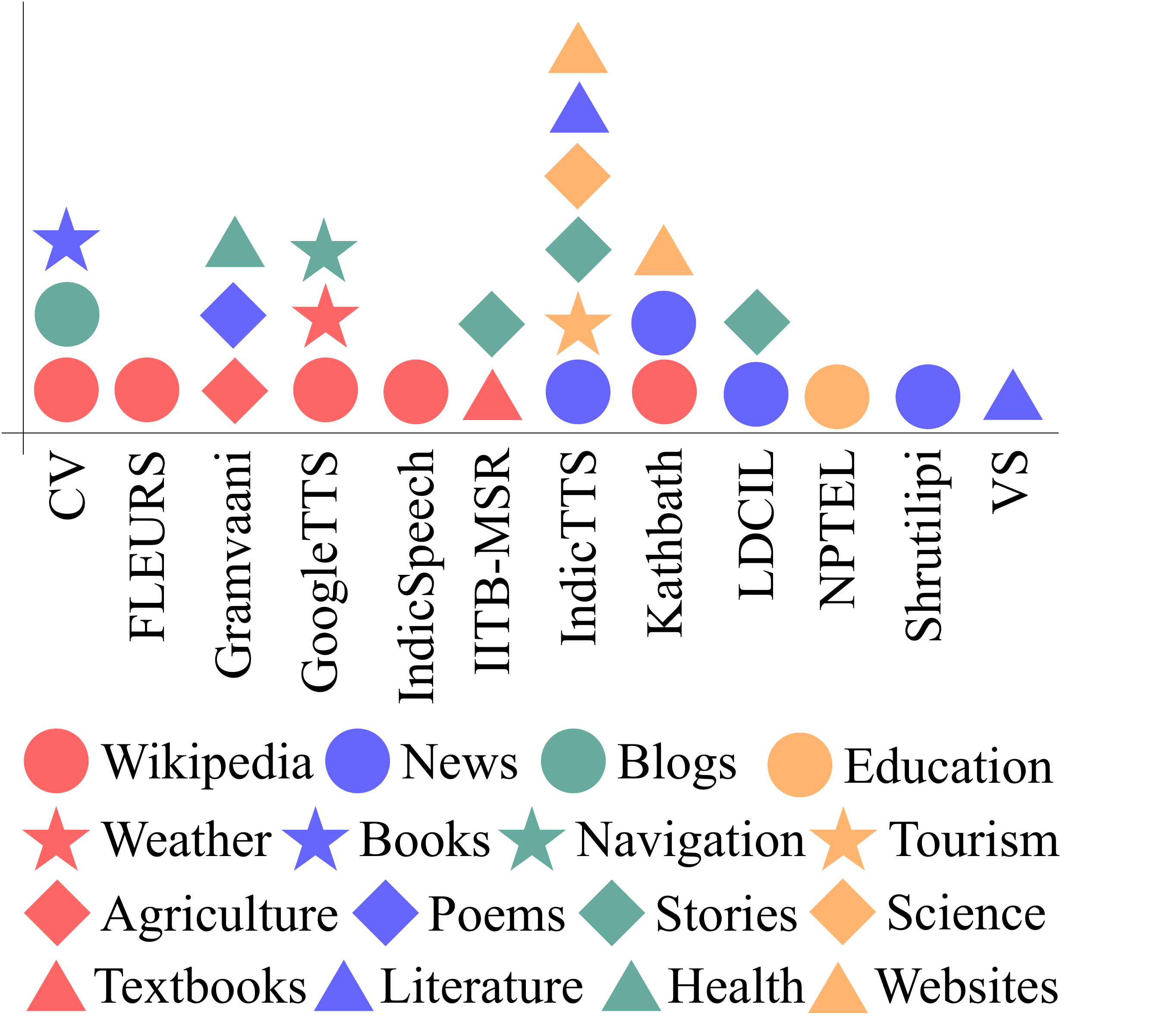}
    \caption{The diversity domains/types covered by the Vistaar benchmark. (CV: CommonVoice;  VS: Vakyasancayah)}
    \label{fig:vistaar}
\end{figure}

To address some of these challenges, we propose collation of benchmarks across languages and domains/types of data.
We call this Vistaar (meaning broad in Hindi) and it comprises of publicly available benchmarks across 12 languages, leading to 59 computed WER values across benchmarks and languages.
We show in Figure~\ref{fig:vistaar} the diversity of the languages and domains/types covered by these benchmarks. 
We will host these datasets in a public resource and release source-code to evaluate an ASR system across these benchmarks.

We then evaluate various ASR systems on Vistaar. 
Specifically, we consider 3 publicly released ASR systems and 2 commercial systems from Google and Microsoft. 
We report the results across models and find large variance across benchmarks, languages, and models.
Specifically, we show that comparing models based on one benchmark may be highly uncorelated with results on another benchmark.
We hope that the Vistaar benchmark, and easy access to it, enables robust evaluation of ASR systems.

\begin{figure}[t!]
    \centering
    \includegraphics[width=\columnwidth]{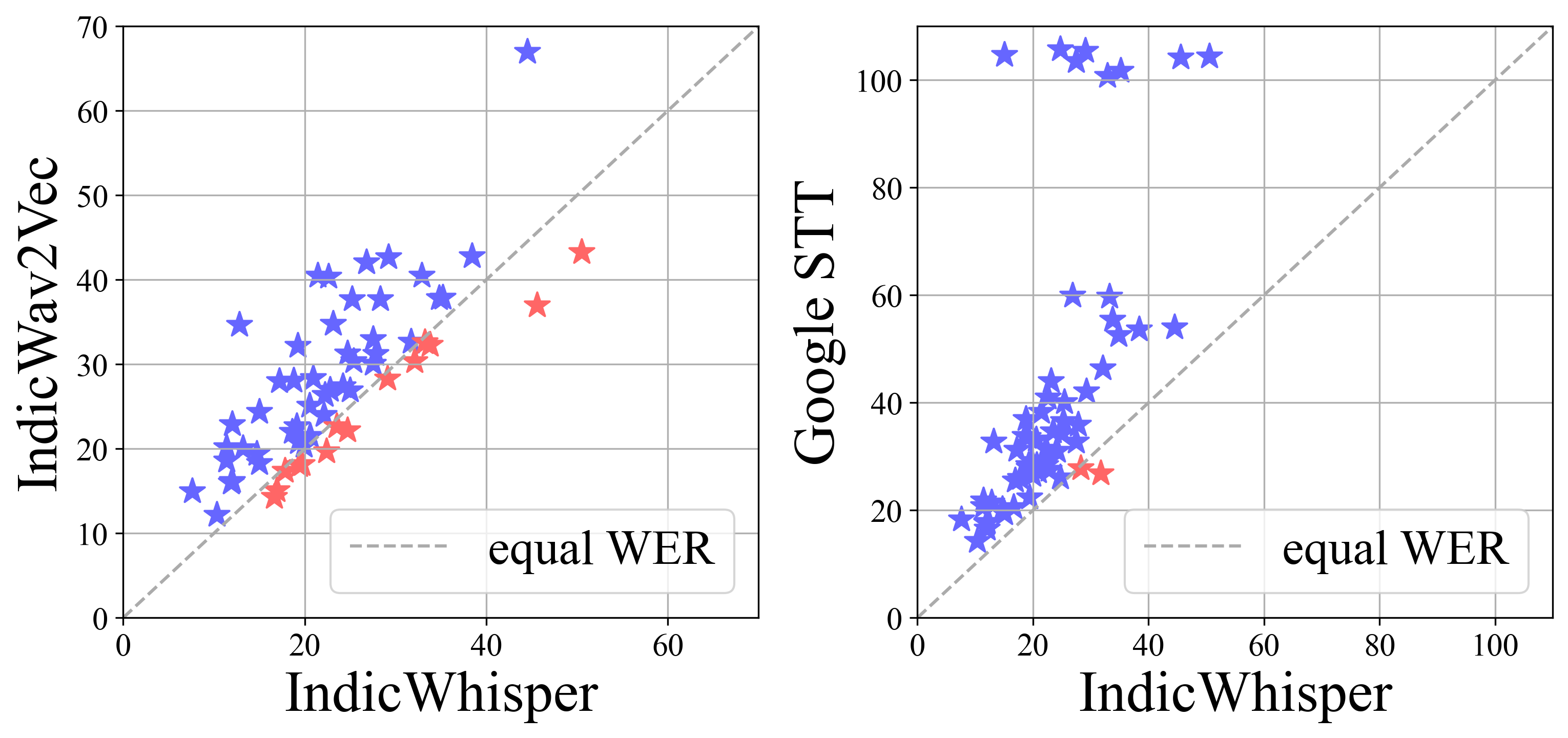}
    \caption{\textbf{WER Comparison on Vistaar:} IndicWhisper is better than IndicWav2Vec in 45 out of 59 benchmarks (highlighted by purple) and better than Google STT in 57 out of 59 benchmarks.}
    \label{fig:vistaar-evaluation}
\end{figure}

Based on our insight that ASR systems' performance varies across domains, we recognize value in training ASR systems with a wide set of training sets. 
To this end, we curate all publicly available training sets for 12 Indian languages amounting over 10,700 hours of audio. 
We call this the Vistaar-train dataset. 
We train the Whisper ASR model \cite{whisper} with Vistaar-train for each of the 12 languages, to create a family of IndicWhisper models.
When evaluated on Vistaar benchmark, we find that IndicWhisper models have the lowest average WER across benchmarks for each of the 12 languages.
Indeed, IndicWhipser has the lowest WER in 39 out of the 59 benchmarks with an average reduction of 4.1 WER. 
We specifically compare IndicWhisper to one publicly available (IndicWav2Vec) and one commercial (Google) ASR system in Figure~\ref{fig:vistaar-evaluation}.
As shown, the reported WER is lower in IndicWhisper for most of the benchmarks, with large gaps for some of the benchmarks.
This validates our proposition that to both evaluate and train ASR systems we need diversity in datasets. 

The rest of the paper is organized as follows.
We detail all benchmark datasets in Vistaar in Section~\ref{sec:vistaar}.
In Section~\ref{sec:vistaar-train} we detail the training dataset and the procedure used to train IndicWhisper.
We compare various public and commercial ASR systems along with IndicWhisper on Vistaar in Section~\ref{sec:evaluation}.
We end with conclusions and our thought-process on further improvements towards building open Indic ASR systems.


\section{Vistaar Benchmark Set}
\label{sec:vistaar}

\begin{table*}[ht!]
\centering
\small
\caption{The Vistaar Benchmark Set - 59 benchmarks over 12 Indian languages }
\label{tab:vistaar-evaluation}
\begin{tabular}{lcccccccccccc}
\toprule
Datasets    & bn                   & gu                   & hi   & kn                   & ml                   & mr                   & or                   & pa                   & \multicolumn{1}{c}{sa}   & ta                   & te                   & ur                   \\
\midrule
Kathbath    & \CheckedBox                 & \CheckedBox                 & \CheckedBox & \CheckedBox                 & \CheckedBox                 & \CheckedBox                 & \CheckedBox                 & \CheckedBox                 & \multicolumn{1}{c}{\CheckedBox} & \CheckedBox                 & \CheckedBox                 & \CheckedBox                 \\
Kathbath-Hard    & \CheckedBox                 & \CheckedBox                 & \CheckedBox & \CheckedBox                 & \CheckedBox                 & \CheckedBox                 & \CheckedBox                 & \CheckedBox                 & \multicolumn{1}{c}{\CheckedBox} & \CheckedBox                 & \CheckedBox                 & \CheckedBox                 \\
FLEURS      & \CheckedBox                 & \CheckedBox                 & \CheckedBox & \CheckedBox                 & \CheckedBox                 & \CheckedBox                 & \CheckedBox                 & \CheckedBox                 &                          & \CheckedBox                 & \CheckedBox                 & \CheckedBox                 \\
CommonVoice & \CheckedBox                 & \multicolumn{1}{l}{} & \CheckedBox & \multicolumn{1}{l}{} & \CheckedBox                 & \CheckedBox                 & \CheckedBox                 & \CheckedBox                 &                          & \CheckedBox                 & \multicolumn{1}{l}{} & \CheckedBox                 \\
IndicTTS    & \CheckedBox                 & \CheckedBox                 & \CheckedBox & \CheckedBox                 & \CheckedBox                 & \CheckedBox                 & \CheckedBox                 & \multicolumn{1}{l}{} &                          & \CheckedBox                 & \CheckedBox                 & \multicolumn{1}{l}{} \\
MUCS        & \multicolumn{1}{l}{} & \CheckedBox                 & \CheckedBox & \multicolumn{1}{l}{} & \multicolumn{1}{l}{} & \CheckedBox                 & \CheckedBox                 & \multicolumn{1}{l}{} &                          & \CheckedBox                 & \CheckedBox                 & \multicolumn{1}{l}{} \\
Gramvaani   & \multicolumn{1}{l}{} & \multicolumn{1}{l}{} & \CheckedBox & \multicolumn{1}{l}{} & \multicolumn{1}{l}{} & \multicolumn{1}{l}{} & \multicolumn{1}{l}{} & \multicolumn{1}{l}{} &                          & \multicolumn{1}{l}{} & \multicolumn{1}{l}{} & \multicolumn{1}{l}{} \\
\bottomrule
\end{tabular}
\end{table*}

In the following we describe various publicly available datasets which we curate as part of the Vistaar benchmark set. 

\noindent \textbf{Kathbath \cite{kathbath}} This contains \textit{read speech} collected on Android phones with the Karya application \cite{iitb_msr} by AI4Bharat, where the sentences are sourced from Wikipedia and news articles. 
This covers all the 12 languages we consider. 

\noindent \textbf{Kathbath-Hard \cite{kathbath}} We create a hard benchmark for  by adding background noise of various types to the audio files of the Kathbath benchmark.
We use ESC dataset\cite{esc}, which consists of 2,000 short clips of background noise from 5 different categories. 
For each audio, we randomly pick a background clip and add it to the audio signal with a random Signal-to-Noise Ratio (SNR) value between 3 dB and 30 dB. 

\noindent \textbf{FLEURS \cite{fleurs}} This contains \textit{read speech of translated Wikipedia content} with 3 recordings by different speakers for a sentence and manual validation collected by researchers at Meta and their collaborators.
This has a large language coverage with 11 of the 12 languages considered, excluding Sanskrit.

\noindent \textbf{CommonVoice \cite{commonvoice}} This contains \textit{crowd-sourced read speech} from the popular Commonvoice website of Mozilla Foundation. 
The exact source of the sentences is not documented but it seems to contain sentences from news, Wiki articles, stories, and literature. 
This covers 8 languages.

\noindent \textbf{IndicTTS \cite{indictts}} This contains \textit{studio-quality read speech} by professional speakers who are trained to maintain constant pitch and prevent stress phenomenon.
Content is sourced from newspapers, websites, blogs, children stories, and tourism and is chosen to consider common day-to-day usage and syllable coverage.
This covers 9 languages. 

\noindent \textbf{MUCS \cite{mucs}} This contains \textit{read speech} collected by different speakers from high-literacy and semi-literate group, and the text is collected from storybooks.
This covers 6 languages.

\noindent \textbf{GramVaani \cite{gramvaani}} This contains \textit{telephone quality speech data} with specific focus on regional/dialectical variations of Hindi using the Uliza crowdsourcing platform by the farmer-centric NGO GramVaani.

The above benchmarks datasets represent a wide variety. 
The variation is across several axes - source content, speakers, audio equipment, and collection agencies.
And the diversity of the benchmarks is evident from a statistic that we obtained when we compared 14 ASR systems on these benchmarks. 
The Spearman correlation of the 14 WER values of the 14 ASR systems on the benchmarks were computed pair-wise.
Certain pairs of benchmarks showed negative correlation. 
For instance, CommonVoice and IndicTTS had a negative correlation of -0.26 indicating that the ranking of models on CommonVoice is not informative of ranking of models on IndicTTS.
This low correlation is perhaps attributable to the diversity in content and speakers of the two benchmarks - while CommonVoice is crowdsources, IndicTTS is collected from professional speakers in a studio environment.

\section{Vistaar-Train Dataset and IndicWhisper Model Training}
\label{sec:vistaar-train}

\begin{table*}[]
\centering
\small
\caption{The Vistaar-Train Dataset With Hours of data across Datasets and Languages}
\label{tab:vistaar_train}
\begin{tabular}{lcccccccccccc|c}
\toprule
Datasets          & bn   & gu  & hi   & kn   & ml  & mr   & or  & pa  & sa  & ta   & te  & ur  & Total \\
\midrule
Shrutilipi        & 443 & 460 & 1620 & 459  & 359 & 1015 & 601 & 94  & 27  & 794  & 390 & 193 & 6457  \\
Kathbath          & 103 & 116 & 137  & 153  & 134 & 172  & 99  & 124 & 102 & 172  & 142 & 74  & 1527  \\
CommonVoice       & 64  & -   & 13   & -    & 1   & 19   & 2   & 2   & -   & 226  & -   & 46  & 373   \\
NPTEL             & 44  & 59  & 157  & 80   & 60  & 68   & -   & -   & -   & 206  & 183 & -   & 857   \\
IISC-mile         & -   & -   & -    & 347  & -   & -    & -   & -   & -   & 150  & -   & -   & 497   \\
MUCS              & -   & 40  & 95   & -    & -   & 94   & 95  & -   & -   & 40   & 40  & -   & 403   \\
IndicTTS          & 20  & 21  & 20   & 19   & 18  & 19   & 19  & -   & -   & 20   & 36  & -   & 192   \\
IITB-msr          & -   & -   & -    & -    & -   & 109  & -   & -   & -   & -    & -   & -   & 109   \\
Gramvaani         & -   & -   & 100  & -    & -   & -    & -   & -   & -   & -    & -   & -   & 100   \\
FLEURS            & 11  & 9   & 7    & 8    & 10  & 12   & 3   & 6   & -   & 9    & 8   & 7   & 90    \\
Vakysancayah      & -   & -   & -    & -    & -   & -    & -   & -   & 78  & -    & -   & -   & 78    \\
Google TTS        & 5   & 8   & -    & 8    & 6   & 3    & -   & -   & -   & 7    & 6   & -   & 43    \\
IIITH-IndicSpeech & 2   & -   & 1    & 2    & 2   & 2    & -   & -   & -   & 1    & 2   & -   & 12    \\
\midrule
Total             & 691 & 712 & 2150 & 1077 & 590 & 1513 & 819 & 226 & 207 & 1625 & 806 & 320 & 10736 \\
\bottomrule
\end{tabular}
\end{table*}

\begin{table*}[]
\centering
\small
\caption{Comparison of publicly-available models on the Hindi subset of the Vistaar benchmark}
\label{tab:vistaar_hindi}
\begingroup
\begin{tabular}{lcccccccc}
\toprule
Model         & Kathbath   & Kathbath-Hard & FLEURS & CommonVoice & IndicTTS & MUCS & Gramvaani & Average  \\
\midrule
Google STT        & 14.3 & 16.7    & 19.4   & 20.8 & 18.3     & 17.8 & 59.9      & 23.9 \\
IndicWav2vec  & 12.2 & 16.2    & 18.3   & 20.2 & 15.0     & 22.9 & 42.1      & 21.0 \\
Azure STT        & 13.6 & 15.1    & 24.3   & 14.6 & 15.2     & 15.1 & 42.3      & 20.0 \\
Nvidia-medium & 14.0 & 15.6    & 19.4   & 20.4 & 12.3     & 12.4 & 41.3      & 19.4 \\
Nvidia-large  & 12.7 & 14.2    & 15.7   & 21.2 & 12.2     & \textbf{11.8} & 42.6      & 18.6 \\
IndicWhisper  & \textbf{10.3} & \textbf{12.0}    & \textbf{11.4}   & \textbf{15.0} & \textbf{7.6 }     & 12.0 & \textbf{26.8 }     & \textbf{13.6} \\

\bottomrule
\end{tabular}
\endgroup
\end{table*}

\begin{table*}[h!]
\centering
\caption{Performance of Publicly available models  on the Vistaar Benchmark}
\label{tab:indicwhisper}
\begin{tabular}{lcccccccccccc|c}
\toprule
Datasets      & bn   & gu   & hi   & kn   & ml   & mr   & or   & pa   & sa   & ta   & te   & ur   & avg  \\
\midrule
\multicolumn{14}{l}{\textbf{Google STT}} \\
\midrule
Kathbath      & 20.6 & 26.0 & 14.3 & 27.9 & 52.6 & 26.6 & 105.7 & 25.5 & 104.2 & 31.1 & 34.3 & 17.6 & 40.5 \\
Kathbath Hard & 22.4 & 28.4 & 16.6 & 31.4 & 53.6 & 28.4 & 105.4 & 29.2 & 104.4 & 32.8 & 35.9 & 20.2 & 42.4 \\
CommonVoice   & 26.1 & -    & 20.8 & -    & 54.0 & 27.5 & 101.7 & 29.2 & -     & 42.1 & -    & 26.9 & 41.0 \\
FLEURS        & 27.3 & 34.7 & 19.4 & 33.8 & 41.0 & 33.2 & 100.8 & 44.0 & -     & 36.6 & 40.1 & 28.4 & 39.9 \\
IndicTTS      & 37.0 & 33.6 & 18.3 & 32.8 & 38.3 & 21.8 & 104.7 & -    & -     & 31.3 & 55.4 & -    & 41.5 \\
MUCS          & -    & 59.7 & 17.8 & -    & -    & 21.4 & 103.5 & -    & -     & 27.8 & 46.4 & -    & 46.1 \\
Gramvaani     & -    & -    & 59.9 & -    & -    & -    & -     & -    & -     & -    & -    & -    & 59.9 \\
\midrule
Average       & 26.7 & 36.5 & 23.9 & 31.5 & 47.9 & 26.5 & 103.6 & 32.0 & 104.3 & 33.6 & 42.4 & 23.3 & 44.3 \\
\midrule
\multicolumn{14}{l}{\textbf{Azure STT}} \\
\midrule
Kathbath      & 16.7 & 22.4 & 13.6 & 23.6 & 43.3 & 20.9 & NA & NA & NA & 28.1 & 22.8 & NA & - \\
Kathbath Hard & 18.4 & 26.7 & 14.9 & 25.9 & 45.7 &      & NA & NA & NA & 30.0 & 24.6 & NA & - \\
CommonVoice   & 15.0 & -    & 14.6 & -    & 47.5 & 19.6 & NA & NA & NA & 39.9 & -    & NA & - \\
FLEURS        & 29.9 & 34.8 & 24.3 & 31.0 & 39.2 & 37.2 & NA & NA & NA & 34.9 & 32.3 & NA & - \\
IndicTTS      & 21.2 & 50.5 & 15.2 & 25.3 & 33.5 & 31.6 & NA & NA & NA & 31.2 & 61.1 & NA & - \\
MUCS          & -    & 23.7 & 15.1 & -    & -    & 27.5 & NA & NA & NA & 25.1 & 16.1 & NA & - \\
Gramvaani     & -    & -    & 42.3 & -    & -    & -    & NA & NA & NA & -    & -    & NA & - \\
\midrule
Average       & 20.2 & 31.6 & 20.0 & 26.4 & 41.8 & 27.4 & NA & NA & NA & 31.5 & 31.4 & NA & - \\
\midrule
\multicolumn{14}{l}{\textbf{IndicWav2Vec}} \\
\midrule
Kathbath      & 14.3 & 17.4 & 12.2 & 20.8 & 37.8 & 20.4 & 22.2 & 15.1 & 37.0 & 27.4 & 26.9 & 16.0 & 22.3 \\
Kathbath Hard & 18.1 & 21.5 & 16.2 & 26.4 & 42.8 & 24.0 & 28.3 & 18.2 & 43.3 & 33.0 & 31.2 & 19.4 & 26.9 \\
FLEURS        & 28.4 & 22.7 & 18.3 & 22.0 & 40.4 & 25.1 & 40.5 & 34.8 & -    & 37.7 & 30.4 & 32.2 & 30.2 \\
CommonVoice   & 31.3 & -    & 20.2 & -    & 67.0 & 27.0 & 37.9 & 19.7 & -    & 42.7 & -    & 32.7 & 34.8 \\
IndicTTS      & 28.1 & 22.7 & 15.0 & 20.1 & 40.5 & 18.6 & 24.4 & -    & -    & 28.0 & 32.3 & -    & 25.5 \\
MUCS          & -    & 32.6 & 22.9 & -    & -    & 34.7 & 30.1 & -    & -    & 37.7 & 30.4 & -    & 31.4 \\
Gramvaani     & -    & -    & 42.1 & -    & -    & -    & -    & -    & -    & -    & -    & -    & 42.1 \\
\midrule
Average       & 24.0 & 23.4 & 21.0 & 22.3 & 45.7 & 25.0 & 30.6 & 21.9 & 40.2 & 34.4 & 30.2 & 25.1 & 28.7 \\
\midrule
\multicolumn{14}{l}{\textbf{IndicWhisper}} \\
\midrule
Kathbath      & 16.6 & 17.8 & 10.1 & 19.3 & 34.8 & 19.9 & 24.7 & 16.9 & 45.6 & 24.2 & 25.0 & 11.9 & 22.2 \\
Kathbath Hard & 19.4 & 20.6 & 11.9 & 22.2 & 38.4 & 22.1 & 29.1 & 19.7 & 50.5 & 27.5 & 27.8 & 14.7 & 25.3 \\
CommonVoice   & 24.7 & -    & 14.2 & -    & 44.5 & 22.8 & 35.2 & 22.4 & -    & 29.2 & -    & 31.7 & 28.1 \\
FLEURS        & 20.9 & 23.5 & 10.5 & 18.6 & 22.6 & 20.5 & 32.9 & 23.1 & -    & 25.2 & 25.4 & 19.2 & 22.0 \\
IndicTTS      & 18.8 & 19.1 & 8.1  & 13.2 & 21.4 & 11.4 & 15.0 & -    & -    & 17.2 & 33.8 & -    & 17.6 \\
MUCS          & -    & 33.2 & 17.1 & -    & -    & 12.8 & 27.5 & -    & -    & 28.3 & 32.1 & -    & 25.2 \\
Gramvaani     & -    & -    & 24.9 & -    & -    & -    & -    & -    & -    & -    & -    & -    & 24.9 \\
\midrule
Average       & 20.1 & 22.8 & 13.8 & 18.3 & 32.3 & 18.2 & 27.4 & 20.5 & 48.0 & 25.3 & 28.8 & 19.4 & 24.6 \\
\bottomrule
\end{tabular}
\end{table*}


In this section, we detail the Vistaar-Train dataset, and then architecture and training methodology to train the IndicWhisper family of models.

\subsection{Vistaar-Train Set}
As discussed in the previous sections, results in correlating the performance of a set of ASR models on a diverse set of benchmarks. 
Based on this, we hypothesize that ASR models must also be trained on a diverse set of training data.
To this end, we first put together Vistaar-Train set which curates 13 publicly released datasets.
In the following, we detail those datasets which are included in Vistaar-Train and not already discussed in the earlier section. 

\noindent \textbf{Shrutilipi \cite{shrutilipi}} This contains \textit{read speech} collected by mining audio and text pairs from news bulletins aired on All India Radio. 
This covers all the 12 languages we consider.

\noindent \textbf{NPTEL} This contains \textit{conversational speech} from classroom lecture recordings of undergraduate and graduate level engineering courses.
The audio files were recorded using a lapel microphone.
This covers 8 languages. 
The recordings are available as videos (most of them are over 30 min long), and the transcripts are available in PDF format. 
We use the recently proposed document-level alignment technique \cite{shrutilipi} to get sentence level transcriptions. We will release this dataset in a public resource. 

\noindent \textbf{IISc-MILE \cite{iisc_mile}} This contains \textit{read speech} recorded in a \textit{clean, noise-free} environment using USB microphones.
Tamil and Kannada data were recorded from 531 and 915 native speakers respectively.

\noindent \textbf{IITB-MSR \cite{iitb_msr}} This contains \textit{read speech} in Marathi recorded from three user groups: (i) low-income rural village, (ii) low-income urban slums, and (iii) university students.
It is collected on Android phones with the Karya \cite{iitb_msr} application, where the sentences are sourced from Marathi textbooks.

\noindent \textbf{Vakyasancayah \cite{vakyasancayah}} This contains \textit{read speech} in Sanskrit collected on Android phones with the Audacity platform.
The text was sourced from pre-classical, classical, and modern Sanskrit literature.

\noindent \textbf{GoogleTTS \cite{googletts, kjartansson-etal-tts-sltu2018}} This contains \textit{read speech} recorded in a \textit{quiet room} with a fanless laptop by Google employees.
Organic sentences were hand-crafted from templates for weather forecasts and navigation.
This covers 7 languages.

\noindent \textbf{IIIT-IndicSpeech \cite{iiit_indicspeech}} This contains \textit{studio-quality read speech}.
A set of 1000 phonetically balanced sentences was selected from the Wikipedia dump of Indian languages released in 2008.
This covers 7 languages.

In combination with the training components of the datasets in Vistaar, the Vistaar-train dataset contains a total of 10,736 hours of data across 12 languages, as detailed in Table~\ref{tab:vistaar_train}.
The size of data varies across languages with Hindi having the maximum of 2,150 hours of data and Sanskrit with 207 hours having the least.
We believe that making this diverse dataset accessible to ASR researchers will speed up research progress.

\subsection{IndicWhisper - Model Architecture and Training}
To demonstrate improved ASR models on the Vistaar-train dataset, we had to make a choice of the model architecture. 
Given the improved performance shown by Whisper models \cite{whisper} released by OpenAI, we chose the pretrained models and fine-tuned them on Vistaar-train. 
We made this choice based on results for Hindi with parts of the training data where Whisper-based models had substantially lower WER than any of the other model architectures. 

We train one model per language and refer to this family as IndicWhisper. 
These models follow Transformer-based encoder-decoder architecture of the Whisper model.
Each model is trained starting from the Whisper-medium model which has 769M parameters and 24 layers in both the encoder and decoder.
It has been trained on 680,000 hours of multilingual data \cite{whisper}.
However, the representation of Indian languages in this data is low, and consequently the performance of the model on Indian languages is significantly poor \cite{whisper}.

We fine-tune the Whisper-medium model for each of the 12 languages using Vistaar-train.
The audio files are re-sampled to 16KHz, mono-channel format, and an 80-channel log-magnitude Mel spectrogram representation is computed with a window size of 25ms and stride of 10ms. 
We use the same byte-level BPE multilingual text tokenizer used in Whisper\cite{whisper}, as it supports all the 12 languages that we consider.
The multilingual tokenizer makes it possible to fine-tune on languages like Odia, although the Whisper\cite{whisper} ASR model does not support it.


Whisper uses a multitask format to perform multiple speech processing tasks like multilingual speech recognition, speech translation, spoken language identification and voice activity detection. 
Specifically, it uses a sequence of input tokens to the decoder for specifying the language (eg. \texttt{<|hi|>}), the task (eg. \texttt{<|transcription|>}, \texttt{<|translation|>}) and whether to predict timestamps or not (\texttt{<|notimestamps|>}).
Since the IndicWhisper model is trained specifically for speech recognition, we pass the \texttt{<|transcription|>} and \texttt{<|notimestamps|>} tokens along with the language token to force the decoder to predict the correct language.


\section{Evaluation Results}
\label{sec:evaluation}
In this section, we evaluate and compare ASR systems on the Vistaar benchmark.

\subsection{ASR systems compared}
We compare the following 6 ASR systems - 3 are publicly available, 2 are commercial systems, and IndicWhisper that is proposed in this paper. 
We detail the 5 ASR systems, excepting IndicWhisper, in the following.

\noindent \textbf{IndicWav2Vec \cite{kathbath}} - The IndicWav2Vec models are Wav2vec Large models with 317M parameters. 
The models are trained on the Kathbath \cite{kathbath} dataset and support all 12 languages.

\noindent \textbf{Nvidia-medium \cite{nemo}} It is a Conformer\cite{conformer} Medium model with 30M parameters, trained on $\approx$1900 hours of Hindi speech, thus supporting only Hindi.

\noindent \textbf{Nvidia-large \cite{nemo}} It is a Conformer\cite{conformer} Large model with 120M parameters, trained on about 2900 hours of speech consisting of Hindi and English, thus supporting only Hindi.

\noindent \textbf{Google STT\footnote{https://cloud.google.com/speech-to-text/}} It is a paid API service with a cost of \$1.44/hour, supporting all 12 languages.

\noindent \textbf{Azure STT\footnote{https://azure.microsoft.com/en-us/products/cognitive-services/speech-to-text}} It is a paid API service with a cost of \$1/hour, supporting all languages except Odia, Punjabi, Sanskrit and Urdu.

\subsection{WER Results}
We report the WER results for each of the 6 ASR systems for the different benchmarks in Vistaar for Hindi in Table~\ref{tab:vistaar_hindi}.
On 6 of the 7 benchmarks for Hindi, IndicWhisper has the lowest WER.
And on average, IndicWhisper has the lowest WER by a significant margin of 5 WER points over the Nvidia-large model.
This is a significant improvement on WER values for the two commercial systems - Google (23.9) and Azure (20). 
Many of the models do not support all 12 languages covered under Vistaar.
For instance, Azure model does not support Odia, Punjabi, Sanskrit and Urdu, while Nvidia models are only available for Hindi. 
We report the the performance of publicly available models on all benchmarks in Table~\ref{tab:indicwhisper}. We show a comparison with IndicWav2Vec and Google in Figure \ref{fig:vistaar-evaluation}.
The WER across languages shows a large variation going from 13.6 in Hindi to 48 in Sanskrit. 
This indicates room for improvement by collecting larger datasets for languages with smaller resources such as Sanskrit, Malayalam, and Odia.
On this entire set of 59 benchmarks across languages, IndicWhisper has the lowest WER for 39, establishing a highly competitive benchmark for Indian language ASR.

\subsection{Discussion}
Training IndicWhisper on diverse training sets clearly improves on all compared ASR systems.
The improvement covers a broad range of benchmarks and languages. 
Indeed, the large gap to commercially available APIs for speech recognition was surprising. 
This suggests that fine-tuning on Whisper-like encoder-decoder architectures trained on large amounts of weakly supervised data will likely bring large improvements in ASR systems for various languages. 
However, given the still large WERs for various languages, we see significant work required for improving Indian language ASR, which may include the following: 
\begin{enumerate}
    \item Curation of large (order tens of thousands of hours) corpora of audio for weakly supervised training.
    \item Building generic acoustic models trained on 1.~which are then combined with domain-specialized language models.
    \item Creation of benchmarks with diversity across speakers, content, and collection methodologies, to evaluate models in 2.
\end{enumerate}

\section{Conclusion}
\label{sec:conclusion}
We made the case that advancing IndicASR requires evaluation of different ASR systems on a diverse set of benchmarks covering languages and types/domains of data. 
The Vistaar benchmark was presented and used to compare various ASR systems.
We also present the IndicWhisper models by finetuning OpenAI's Whisper models on the Vistaar-train set with over 10,000 hours on 12 Indian languages.
IndicWhisper achieves significantly lower WER across a large set of benchmarks establishing state-of-the-art performance.
However, the obtained WER results indicate further room for improvement, and we outlined potential directions of research.

\section{Acknowledgements}

We would like to thank the Ministry of Electronics and Information Technology (MeitY\footnote{https://www.meity.gov.in/}) of the Government of India and the Centre for Development of Advanced Computing (C-DAC\footnote{https://www.cdac.in/index.aspx?id=pune}), Pune for generously supporting this work and providing us access to multiple GPU nodes on the Param Siddhi Supercomputer. We would like to thank the EkStep Foundation and Nilekani Philanthropies for their generous grant which went into hiring human resources as well as cloud resources needed for this work. We would like to thank Megh Makhwana from Nvidia for helping in training Conformer-based ASR models.

\bibliographystyle{IEEEtran}
\bibliography{mybib}

\end{document}